\documentclass[10pt,journal,compsoc]{IEEEtran}
\ifCLASSOPTIONcompsoc
  \usepackage[nocompress]{cite}
\else
  \usepackage{cite}
\fi
\ifCLASSINFOpdf
\else
\fi

\usepackage{amsmath,amsfonts,bm}

\def\eqref#1{equation~\ref{#1}}

\def\1{\bm{1}}

\DeclareMathAlphabet{\mathsfit}{\encodingdefault}{\sfdefault}{m}{sl}
\SetMathAlphabet{\mathsfit}{bold}{\encodingdefault}{\sfdefault}{bx}{n}

\usepackage{hyperref}
\usepackage{url}
\usepackage{graphicx}
\usepackage{algorithm}
\usepackage{algpseudocode}
\usepackage{multirow}

\begin{document}
%
\title{Training Data Subset Search with \\Ensemble Active Learning}
%
%
%
%

\author{Kashyap Chitta,
        Jos\'{e} M. \'{A}lvarez,
        Elmar Haussmann,
        and Cl\'{e}ment Farabet
\IEEEcompsocitemizethanks{\IEEEcompsocthanksitem K. Chitta is with the Max Planck Institute for Intelligent Systems, T\"{u}bingen and University of T\"{u}bingen. Work completed during an internship at NVIDIA.\protect\\
E-mail: kashyap.chitta@tue.mpg.de
\IEEEcompsocthanksitem J. M. \'{A}lvarez, E. Haussmann and C. Farabet are with NVIDIA.}
}

%
%

\markboth{}%
{Chitta \MakeLowercase{\textit{et al.}}: Training Data Subset Search with Ensemble Active Learning}
%



\IEEEtitleabstractindextext{%
\begin{abstract}
Deep Neural Networks (DNNs) often rely on very large datasets for training. Given the large size of such datasets, it is conceivable that they contain certain samples that either do not contribute or negatively impact the DNN's optimization. Modifying the training distribution in a way that excludes such samples could provide an effective solution to both improve performance and reduce training time. In this paper, we propose to scale up ensemble Active Learning (AL) methods to perform acquisition at a large scale (10k to 500k samples at a time). We do this with ensembles of hundreds of models, obtained at a minimal computational cost by reusing intermediate training checkpoints. This allows us to automatically and efficiently perform a training data subset search for large labeled datasets. We observe that our approach obtains favorable subsets of training data, which can be used to train more accurate DNNs than training with the entire dataset. We perform an extensive experimental study of this phenomenon on three image classification benchmarks (CIFAR-10, CIFAR-100 and ImageNet), as well as an internal object detection benchmark for prototyping perception models for autonomous driving. Unlike existing studies, our experiments on object detection are at the scale required for production-ready autonomous driving systems. We provide insights on the impact of different initialization schemes, acquisition functions and ensemble configurations at this scale. Our results provide strong empirical evidence that optimizing the training data distribution can provide significant benefits on large scale vision tasks.
\end{abstract}

}

\maketitle
\IEEEdisplaynontitleabstractindextext

%
\IEEEpeerreviewmaketitle

\IEEEraisesectionheading{\section{Introduction}\label{sec:introduction}}

Deep Neural Networks (DNNs) have become the dominant approach for addressing supervised learning tasks involving high-dimensional inputs. There is significant interest in automating the end-to-end process of applying DNNs to real-world problems such as training perception systems for autonomous driving \cite{Wong2018Transfer,He2018AMC,He2019AutoML}. While there has been a considerable effort towards methods and frameworks that automate DNN architecture search \cite{Elsken2018Neural,Luo2018Neural,Liu2018DARTS,Xie2018SNAS,Dong2019Searching} and training hyper-parameter search \cite{Golovin2017Google,Mutny2018Efficient,Kandasamy2019Tuning}; the process of searching for the right training data distribution (also called dataset curation) is still performed by experts, requiring several heuristics and significant manual effort. With the rapid growth in the availability of labeled data for perception tasks, to the order of billions of samples \cite{Sun2017Revisiting,Yalniz2019Billion}, automating the \textit{training data subset search} would make the application of DNNs much easier for non-experts, and potentially lead to datasets and models that outperform those that were curated by hand.

In this paper, we present a simple yet effective method to perform a training data subset search by using {ensemble Active Learning (AL)}. The typical goal of AL is to select, from a large unlabeled dataset, the smallest possible training set to label in order to solve a specific task \cite{Cohn1994Active}. We instead propose to use AL to build \textit{data subsets} of a large labeled training dataset that give more accurate DNNs in less training time. We demonstrate that this approach can automatically curate large datasets. We study the impact of key design choices for AL, and the robustness of the selected subsets to changes in model architectures.

We tackle two key issues that have not been addressed so far by state-of-the-art AL methods. The first is the difficulty in \textbf{scaling the number of models} for the popular \textit{ensemble AL} technique. While it seems intuitive that more ensembles can improve performance, existing studies show no gains in AL performance beyond 10 models, and even {recommend} the use of only 5 models \cite{Lakshminarayanan2017Simple,Ovadia2019Trust}. In this study, we propose the use of implicit ensembles with {hundreds of training checkpoints} from different experimental runs, and empirically demonstrate the effectiveness of this approach.

Second, we switch to a \textbf{large-scale experimental setting} compared to what is typically used for AL experiments. For example, Beluch et al. \cite{Beluch2018Power} and Sinha et al. \cite{Sinha2019Variational} never use more than 30\% of the ImageNet dataset, and do not compete with the full dataset performance. In contrast, for ResNet-18 training with 80\% of ImageNet, we improve the top-1 accuracy by 0.5\% over a model trained with the entire ImageNet dataset. For object detection, existing studies are limited to acquisition at the order of ~10k samples, beyond which improvements become marginal \cite{Kao2018Localization,Brust2019Active,Desai2019Adaptive,Aghdam2019Active}.  In this paper, we scale the process to the acquisition of 200k images in each iteration, where the final selected subset outperforms training with all the available data. Our experiments are the first to provide insights regarding automatic dataset curation at the scale required for production-ready autonomous driving systems.

To summarize, our contributions are as follows: (1) we propose a simple approach to scale up ensemble AL methods to hundreds of models with a negligible computational overhead at train time, and (2) we conduct a detailed empirical study on how to effectively {reduce the size of existing datasets}, containing millions of samples, without human curation. Our study provides practical insights by covering several large datasets for object detection and image classification.

\section{Related Work}

A comprehensive review of classical approaches to AL is presented in \cite{Settles2010Active}. In these approaches, uncertainty is used as a criteria for selecting which unlabeled samples should be annotated. While some methods focus on information-theoretic measures of uncertainty \cite{Gal2015Dropout,Gal2015Bayesian,Gal2017Deep,Gal2018Sufficient,Smith2018Understanding}, others use learned scores aimed at measuring information gain \cite{Yoo2019Learning,Sinha2019Variational}. Among these methods, ensemble-based AL techniques are of particular note, due to their conceptual simplicity and state-of-the-art results \cite{Beluch2018Power,Chitta2018Deep,Chitta2018Large}. In these methods, a measure of variance in the outputs of an ensemble of differently initialized DNNs is used as an uncertainty measure for selecting samples to label. Ensembles of neural networks have strong performance \cite{Hansen1990Neural,Dietterich2000Ensemble}, as well as good scalability and robustness to dataset shifts, making them ideal for large-scale AL applications \cite{Lakshminarayanan2017Simple,Ovadia2019Trust}. 

There are few existing works on AL for object detection with DNNs. Roy et al. \cite{Roy2018Deep} use a query by committee approach and the disagreement between the convolutional layers in the object detector backbone to query images. Brust et al. \cite{Brust2019Active} evaluate a set of uncertainty-based AL aggregation metrics that are suitable for most object detectors. These metrics are focused on the aggregation of the scores associated to the detected bounding boxes. Kao et al. \cite{Kao2018Localization} propose an algorithm that incorporates the uncertainty of both classification and localization outputs to measure how tight the detected bounding boxes are. This approach computes two forward passes, the original image and a noisy version of it, and compares how stable the predictions are. Desai et al. \cite{Desai2019Adaptive} combine AL with weakly supervised learning to reduce the efforts needed for labeling. Instead of always querying accurate bounding box annotations, their method first queries a weak annotation consisting of a rough labeling of the object center, and move towards the accurate bounding box annotation only when required. They show promising results in terms of the amount of time saved for annotation. More recently, Aghdam et al. \cite{Aghdam2019Active} propose an image-level scoring process to rank unlabeled images for their automatic selection. They first compute the importance of each pixel in the image and aggregate these pixel-level scores to obtain a single image-level score. 

All these methods show promising results using different object detectors on relatively smaller datasets such as PASCAL VOC \cite{Everingham2010Pascal} and MS-COCO \cite{Lin2014Microsoft}. However, their experiments are focused on the early stage of the AL process, and therefore consider only small dataset sizes (from 500 up to 3500 images in the case of PASCAL VOC). In general, once the number of training images increases, the improvement of those approaches becomes marginal, as it occurs for instance in~\cite{Kao2018Localization} with MS-COCO where the training set size reaches 9000 images. None of these methods provide insights into the applicability of AL for dataset curation in the object detection setting.

It is occasionally observed in the classical AL literature that training on a subset of data can give better models than training on the full dataset \cite{Schohn2000Less,Lapedriza2013Training,Gavves2015Active,Konyushkova2015Introducing}. However, this is not investigated with state-of-the-art AL techniques for DNNs. Contemporary techniques typically assume that the performance obtained by training on the entire data pool is an upper bound. We show that the full dataset is not an upper bound, demonstrating distinct advantages of training on data subsets when they are sufficiently large. 

Despite the widespread use of image classification datasets, there has only recently been an increased interest in understanding the properties of subsets of these datasets. \textit{Core-set selection} \cite{Sener2017Active} is one such attempt to reduce dataset sizes, by finding a representative subset of points based on relative distances in the DNN feature space. Vodrahalli et al. \cite{Vodrahalli2018Training} also aims to find representative subsets, using the magnitude of the gradient generated by each sample for the DNN as an importance measure. It is important to note that these techniques are {unable to match or improve} the performance of a model trained with all the data for DNNs. More recent techniques are able to successfully reduce dataset sizes by identifying redundant examples, albeit to a much smaller extent than our approach. Birodkar et al. \cite{Birodkar2019Semantic} uses clustering in the DNN feature space to identify redundant samples, leading to a discovery of 10\% redundancy in the CIFAR-10 and ImageNet datasets. \textit{Selection via proxy} \cite{Coleman2019Selection} uses active learning techniques to select a subset of data for training, and removes samples in CIFAR-10 without reducing the network performance. In comparison to these methods, our technique not only maintains, but also improves the performance of a DNN. 

A second category of works on data subsets is closely related to catastrophic forgetting in DNNs. Toneva et al. \cite{Toneva2018Empirical} uses the number of instances in which a previously correctly classified sample is 'forgotten' and misclassified during training as an importance measure. By doing so, this method is able to remove 30\% of the samples that are 'unforgettable' from CIFAR-10 without significantly reducing performance. Chang et al. \cite{Chang2017Active} propose a similar idea of emphasizing data points whose predictions have changed most over the previous training epochs. Rather than directly subsampling, this variance in predictions is used to increase or decrease the sampling weight during training, and therefore the approach has no significant impact on training time. In comparison, by training on only a specific subset, we not only improve performance but also cut down training time by 20\% to 50\% for the datasets used in our study.

\section{Data Subset Search for Classification}

We now show how ensembles can be used for uncertainty estimation for classification DNNs. We then describe an algorithm by which AL can be applied for building data subsets using ensemble uncertainties. We focus on three design choices within this algorithm: the initialization scheme, acquisition function and ensemble configuration.

Consider a distribution ${\tt p}(\textbf{x},y)$ over inputs $\textbf{x}$ and labels $y$. In a Bayesian framework, the \textit{predictive uncertainty} of a particular input $\textbf{x}^{*}$ after training on a dataset $L$ is denoted as ${\tt P}(y=k|\textbf{x}^{*},L)$. The predictive uncertainty will result from \emph{data (aleatoric) uncertainty} and \emph{model (epistemic) uncertainty} \cite{Kendall2017What}. A model's estimates of \emph{data uncertainty} are described by the posterior distribution over class labels $y$ given a set of model parameters $\theta$. This is typically the softmax output in a classification DNN. Additionally, the \emph{model uncertainty} is described by the posterior distribution over the parameters $\theta$ given the training data $L$ \cite{Malinin2018Predictive}:
\begin{equation}
\underbrace{{\tt P}(y=k | \textbf{x}^{*}, L)}_{Predictive} = \int \underbrace{{\tt P}(y=k|\textbf{x}^{*},  {\theta})}_{Data}\underbrace{{\tt p}(\theta | L)}_{Model} d\theta.
\label{eqn:unc}
\end{equation}

We see that, uncertainty in the model parameters ${\tt p}(\theta | L)$ induces a distribution over the softmax distributions ${\tt P}(y=k| \textbf{x}^{*},  { \theta})$. The expectation is obtained by marginalizing out the parameters $\theta$. Unfortunately, obtaining the full posterior ${\tt p}(\theta|L)$ using Bayes' rule is intractable. If we train a single DNN, we only obtain a single sample from the distribution ${\tt p}(\theta | L)$. Ensemble uncertainty estimation techniques approximate the integral from Eq. \ref{eqn:unc} by Monte Carlo estimation, generating multiple samples using different members of an ensemble \cite{Lakshminarayanan2017Simple}:
\begin{equation}
{\tt P}(y=k | \textbf{x}^{*}, L) \approx \frac{1}{E}\sum_{e \in E} {\tt P}(y=k| \textbf{x}^{*},  {\theta}_e ),\  {\theta}_e \sim {\tt q}( {\theta}),
\label{eqn:uncsamp}
\end{equation}

\noindent
where ${\tt q}( {\theta})$ represents an approach used for building the ensemble. The strength of the ensemble approximation depends on both the number of samples drawn ($E$), and how the parameters for each model in the ensemble are sampled (i.e., how closely the distribution ${\tt q}( {\theta})$ matches ${\tt p}(\theta|L)$). In this work, we study in more detail the impact of the number of samples $E$. Our overall algorithm, presented next, is generic and can potentially benefit from more advanced techniques for uncertainty estimation that attempt to better match ${\tt q}( {\theta})$ from Eq. \ref{eqn:unc} and ${\tt p}(\theta|L)$ from Eq. \ref{eqn:uncsamp}, which is an open and active research area \cite{Malinin2018Predictive}.

\subsection{Data Subset Search}

Our approach is a modification of typical AL, and involves the following:

\begin{enumerate}
    \item A \textbf{labeled dataset}, consisting of $N_l$ labeled pairs, $L=\{(\textbf{x}_l^j,y_l^j)\}_{j=1}^{N_l}$, where each $\textbf{x}^j \in X$ is a data point and each $y^j \in Y$ is its corresponding label. 
    \item An \textbf{acquisition model}, $\mathcal{M}_a:X\rightarrow Y$. For our ensemble AL approach, the acquisition model $\mathcal{M}_a$ takes the form of a set of $E$ different DNNs with parameters $\{\theta_a^{(e)}\}_{e=1}^{E}$, initialized based on a particular initialization scheme (Section \ref{sec:init}) and ensemble configuration (Section \ref{sec:ens}).
    \item A \textbf{data subset}, $S=\{(\textbf{x}_s^j,y_s^j)\}_{j=1}^{N_s}$, where $S$ is a subset of $L$ selected using an acquisition function $\alpha(\textbf{x},\mathcal{M}_a)$ (Section \ref{sec:acq}).
    \item A \textbf{subset model}, $\mathcal{M}_s:X\rightarrow Y$, with parameters $\theta_s$, which is the model trained on the selected data subset $S$.
\end{enumerate}

We decouple the data selection and final optimization of AL into two different models. The \textit{acquisition model} selects a subset of data using ensemble uncertainty estimation, which is then used to optimize the parameters of the \textit{subset model}.

\begin{algorithm}[t!]
\caption{\textit{Pretrain} scheme.}
\label{alg:pretrain}
\begin{algorithmic}
    \State {Initialize $S \leftarrow \{\}$}
    \State {Pretrain acquisition ensemble $\mathcal{M}_a$ on $L$}
    \State {Pretrain subset model $\mathcal{M}_s$ on $L$}
    \State Compute acquisition function $\alpha(\textbf{x},\mathcal{M}_a) \ \forall \ \textbf{x}\in L$
    \State Append $N_s$ samples with maximum $\alpha$ to $S$
    \State Use $S$ to fine-tune parameters $\theta_s$ of subset model
\end{algorithmic}
\end{algorithm}

\begin{algorithm}[t!]
\caption{\textit{Compress} scheme.}
\label{alg:compress}
\begin{algorithmic}
    \State {Initialize $S \leftarrow \{\}$}
    \State {Pretrain acquisition ensemble $\mathcal{M}_a$ on $L$}
    \State {Randomly initialize subset model $\mathcal{M}_s$}
    \State Compute acquisition function $\alpha(\textbf{x},\mathcal{M}_a) \ \forall \ \textbf{x}\in L$
    \State Append $N_s$ samples with maximum $\alpha$ to $S$
    \State Use $S$ to optimize parameters $\theta_s$ of subset model
\end{algorithmic}
\end{algorithm}

\begin{algorithm}[t!]
\caption{\textit{Build Up} scheme.}
\label{alg:buildup}
\begin{algorithmic}
    \State {Initialize $S \leftarrow \frac{N_s}{8}$ random samples from $L$}
    \State {Pretrain acquisition ensemble $\mathcal{M}_a$ on $S$}
    \While{$Size(S)<N_s$}
    \State {Randomly initialize subset ensemble $\mathcal{M}_s$}
    \State Compute acquisition function $\alpha(\textbf{x},\mathcal{M}_a) \ \forall \ \textbf{x}\in L$
    \State Move $Size(S)$ samples with maximum $\alpha$ from $L$ to $S$
    \State Use $S$ to optimize parameters $\{\theta_s^{(e)}\}_{e=1}^{E}$ of subset
    \State ensemble
    \EndWhile
\end{algorithmic}
\end{algorithm}

\subsection{Initialization Schemes}
\label{sec:init}

We consider three different \textit{initialization schemes} for the acquisition and subset models: pretrain, compress and build up (Algorithms \ref{alg:pretrain}, \ref{alg:compress} and \ref{alg:buildup}). For the pretrain and compress schemes, the subset $S$ is initialized with an empty set. In the {pretrain} scheme, the entire dataset $L$ is used for pretraining and initializing the parameters of both the acquisition and subset models. During optimization, the subset model is then finetuned on the data subset $S$. In the {compress} scheme, the acquisition model is pretrained on $L$ but the subset model is randomly initialized and trained from scratch on $S$.

Finally, in the build up scheme, we follow an iterative AL loop. Specifically, we start by initializing $S$ with a randomly selected subset of the data and train an acquisition model ensemble. After performing acquisition, instead of training a single subset model, we optimize an ensemble of $E$ subset models. This ensemble is used as an acquisition model for a subsequent iteration on the remaining unselected data. Our goal is to finally reach a subset of $N_s$ samples. As observed by \cite{Chitta2018Large}, exponentially growing the dataset size offers practical benefits in an AL loop setting for classification. We therefore follow this approach, by initializing $S$ with $\frac{N_s}{8}$ random samples, and iterating two further times at $\frac{N_s}{4}$ and $\frac{N_s}{2}$ samples before obtaining a final subset of size $N_s$.

\subsection{Acquisition Functions}
\label{sec:acq}

In our experiments, we empirically evaluate four acquisition functions of the form $\alpha(\textbf{x},\mathcal{M}_a)\rightarrow \mathbb{R}$ for AL-- entropy, mutual information, variation ratios and error count. We choose these well-known acquisition functions to maintain simplicity and scalability. For a detailed theoretical analysis of these acquisition functions, we refer the reader to \cite{Gal2016Uncertainty}.

\noindent \textbf{Entropy:} In the case of classification, the predictive uncertainty for a sample ${\tt P}(y=k | \textbf{x}^{*}, L)$ from Eq. \ref{eqn:unc} is a multinomial distribution, which can be represented as a vector of probabilities $\textbf{p}$ over each of the $K$ classes. We can obtain the predictive uncertainty for a sample as its entropy \cite{Shannon1948Mathematical}:

\begin{equation}
    \mathcal{H}(\textbf{p}) = - \textbf{p}^T \log \textbf{p}.
    \label{eqn:ent}
\end{equation}

\noindent \textbf{Mutual Information:}, also called BALD and Jensen-Shannon Divergence, explicitly looks for large disagreement between the models (i.e., \textit{model uncertainty})~\cite{Houlsby2011Bayesian,Smith2018Understanding}: 

\begin{equation}
    \underbrace{\mathcal{J}(\textbf{p})}_{Model} =  \underbrace{\mathcal{H}(\textbf{p})}_{Predictive} - \underbrace{\frac{1}{E}\sum_{e \in E} \mathcal{H}(\textbf{p}^{(e)})}_{Data},
    \label{eqn:mi}
\end{equation}

\noindent
where $\textbf{p}^{(e)}$ denotes the prediction of an individual member of the ensemble before marginalization. Since entropy is always positive, the maximum possible value for $\mathcal{J}(\textbf{p})$ is $\mathcal{H}(\textbf{p})$. However, when the models make similar predictions, $\frac{1}{E}\sum_{e=1}^E \mathcal{H}(\textbf{p}_e) \rightarrow \mathcal{H}(\textbf{p})$, and $\mathcal{J}(\textbf{p}) \rightarrow 0$, which is its minimum value. This shows that $\mathcal{J}$ encourages samples with high disagreement to be selected during the data acquisition process. An alternate way to look at the metric is that from the predictive uncertainty, we subtract away the expected \textit{data uncertainty}, leaving an approximate of the \textit{model uncertainty} \cite{Depeweg2017Decomposition}.

\noindent \textbf{Variation Ratios:} also looks for disagreement between the models~\cite{Gal2016Uncertainty} and it is defined as the fraction of members in the ensemble that do not agree with the majority vote $M = \underset{e \in E}{\text{Mode}}(\underset{k \in K}{\arg \max \ } \textbf{p}_k^{(e)})$:
\begin{equation}
    \mathcal{V}(\textbf{p}) = 1 - \frac{1}{E}\sum_{e \in E} (\underset{k \in K}{\arg \max \ } \textbf{p}_k^{(e)} = M),
    \label{eqn:vr}
\end{equation}
where $K$ is the number of classes. This is the simplest quantitative measure of variation, and prior applications in literature show that it works well in practice \cite{Gal2017Deep,Beluch2018Power,Chitta2018Adaptive}. 

\noindent \textbf{Error Count:} is similar to variation ratios, but checks for disagreement with the ground truth label $y$ rather than the mode of predictions $M$:
\begin{equation}
    \mathcal{E}(\textbf{p}) = 1 - \frac{1}{E}\sum_{e \in E} (\underset{k \in K}{\arg \max \ } \textbf{p}_k^{(e)} = y).
    \label{eqn:ec}
\end{equation}

This is typically not used in AL experiments as it cannot be computed without the ground truth labels. In our setting, where the labels are available, we use this function as a baseline which, in principle, prioritizes mistakes made by the network when selecting data.
\begin{figure}[t!]
\centering
\includegraphics[width=\columnwidth]{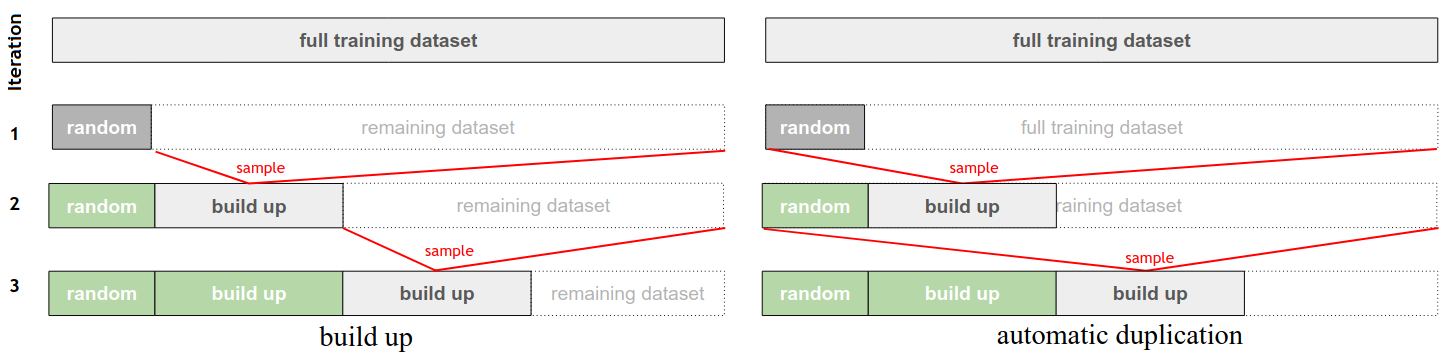}
\vspace{-0.5cm}
\caption{Comparison of build up and automatic duplication initialization schemes for object detection. Instead of sampling from only the remaining dataset, automatic duplication involves sampling from the entire dataset at each iteration, allowing certain repeated samples in the final training set.}
\label{fig:dup}
\vspace{-0.0cm}
\end{figure}
\begin{figure}[!t]
 \begin{tabular}{ccc}
 \hspace{-0.2cm}\includegraphics[width=0.33\linewidth] {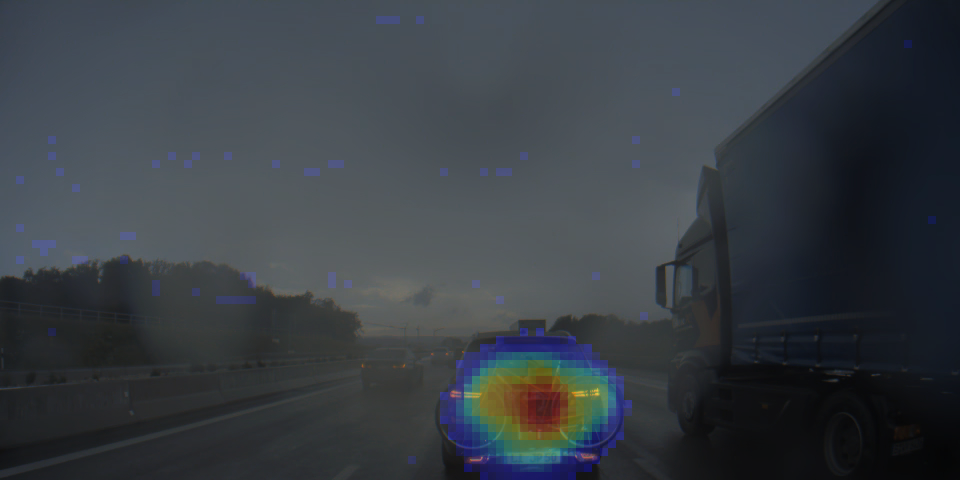}& \hspace{-0.38cm}\includegraphics[width=0.33\linewidth] {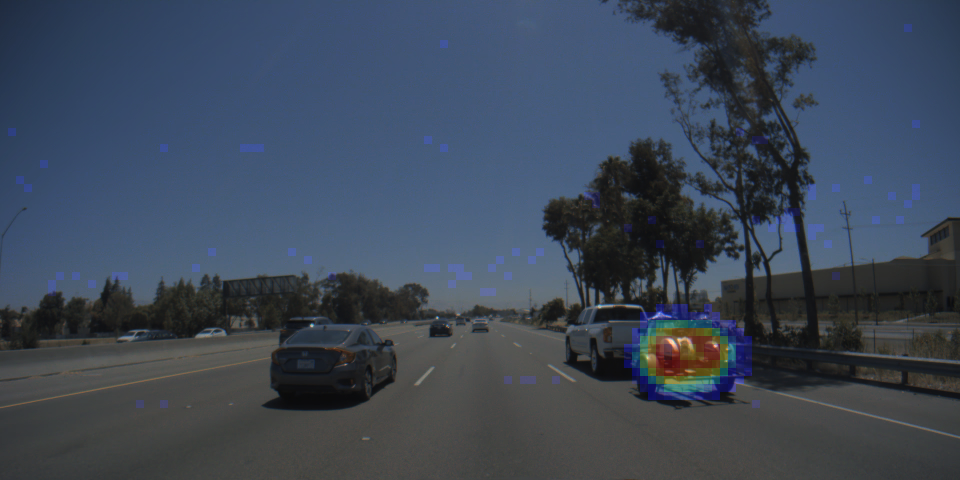}&\hspace{-0.38cm}\includegraphics[width=0.33\linewidth] {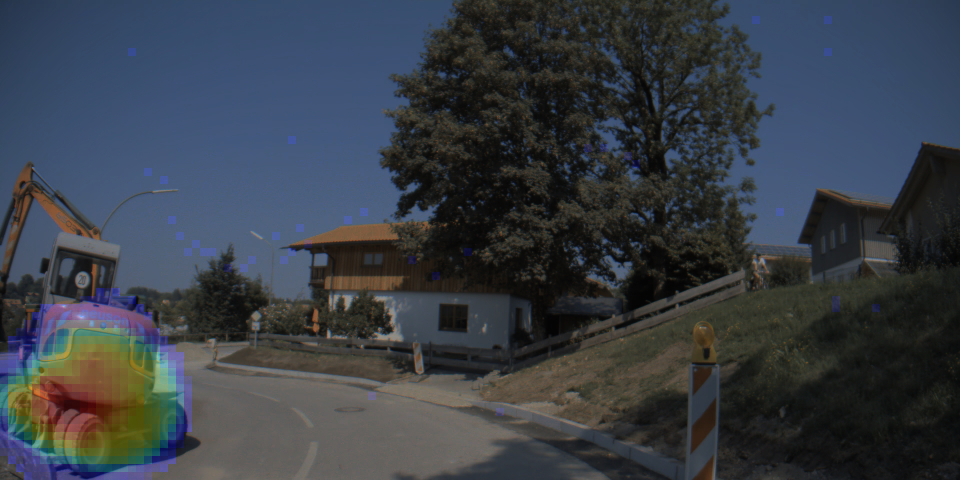}
 \end{tabular}
  \vspace{-0.4cm}
   \caption{Uncertainty heatmaps representing the acquisition function (in this case, mutual information) at each pixel for a given class. These scores are then aggregated to provide a single value per image by taking the maximum over all locations and classes.}
  \label{fig:heatmaps}
  \vspace{-0.3cm}
 \end{figure}
\subsection{Ensemble Configurations}
\label{sec:ens}
State-of-the-art ensemble-based AL approaches use different random seeds to construct ensembles \cite{Beluch2018Power}. They recommend the number of samples drawn to be in the range $E\in(5,10)$ models \cite{Lakshminarayanan2017Simple}. In theory, the error of a Monte Carlo estimator should decrease with more samples, which is evident in other BNN based uncertainty estimation techniques, that require the number of stochastic samples drawn to be increased to the range $E\in(50,100)$ \cite{Gal2015Dropout}.

The major limiting factor preventing the training of $E\in(50,100)$ models with different random seeds for ensemble AL is the computational burden at train time. Implicit ensembling approaches that are computationally inexpensive, such as Dropout \cite{Srivastava2014Dropout}, suffer from mode collapse, where the different members in the ensemble lack sufficient diversity for reliable uncertainty estimation \cite{Pop2018Deep}. An alternate approach, called snapshot ensembles, that is less computationally expensive at train time, uses a cyclical learning rate to converge to multiple local optima in a single training run \cite{Huang2017Snapshot}. However, this technique is also limited to ensembles in the range of $E=6$ members. In our work, we present an implicit ensembling approach that allows users to draw a large number of samples using the catastrophic forgetting property in DNNs \cite{Toneva2018Empirical}. Specifically, we exploit the disagreement between different checkpoints stored during successive training epochs to efficiently construct large and diverse ensembles. We collect several training checkpoints over multiple training runs with different random seeds. This allows us to maximize the number of samples drawn, efficiently generating ensembles with up to hundreds of members.

\begin{figure*}[!t]
\centering
\hspace{-0.45cm}\begin{tabular}{ccc}
\includegraphics[width=0.33\textwidth]{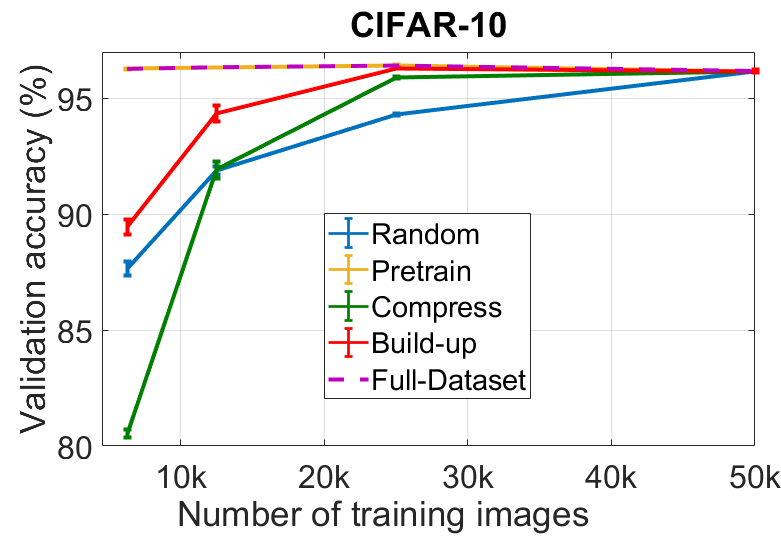}&\hspace{-0.4cm}
\includegraphics[width=0.33\textwidth]{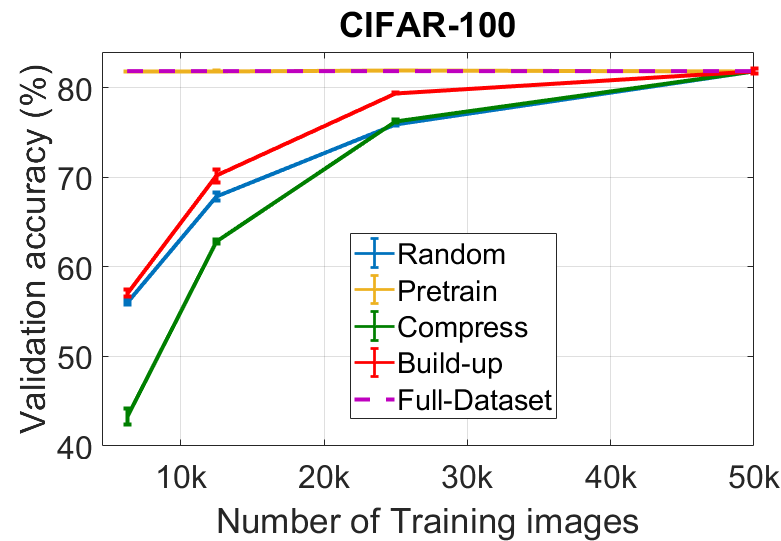}&\hspace{-0.4cm}
\includegraphics[width=0.33\textwidth]{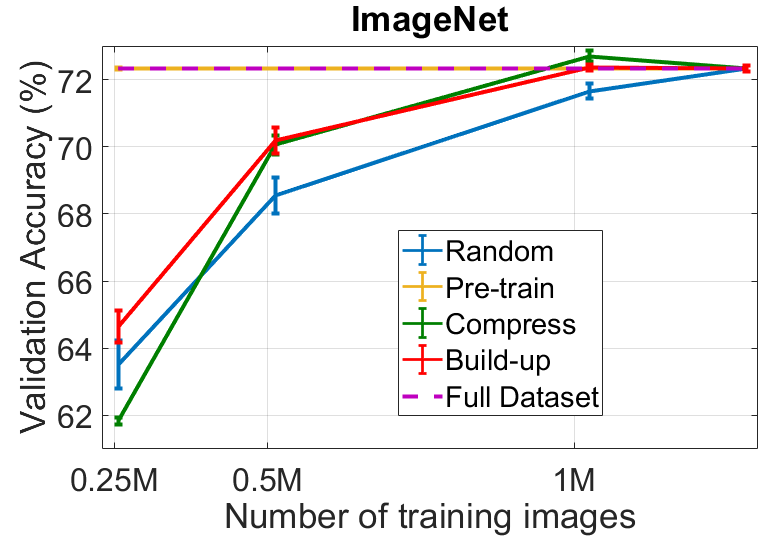}
\end{tabular}
\vspace{-0.35cm}
\caption{Comparison of pretrain, compress and build up initialization schemes on CIFAR-10, CIFAR-100 and ImageNet. Results shown are mean and std over three trials of the top-1 validation accuracy of an ensemble of ResNet-18 models. While compress does not perform well on small subsets of data, build up provides strong results on all subset sizes for all three tasks.}
\label{fig:init}
\vspace{-0.25cm}
\end{figure*}
\begin{table*}[!t]
	\begin{center}
    \caption{Top-1 validation accuracy of ensembles trained on subsets of CIFAR-10, CIFAR-100 and ImageNet (mean of 3 trials), comparing different acquisition functions. Each column indicates a subset with a size equal to the specified percentage of the full dataset. Mutual information ($\mathcal{J}$) and variation ratios ($\mathcal{V}$) give the best performance.}
    \vspace{-0.25cm}
    \setlength{\tabcolsep}{12pt}
    \label{tab:subsample}
		\begin{tabular}{c|c|c|c||c|c|c||c|c|c}
			\hline
			\textbf{Dataset}&\multicolumn{3}{c||}{ \textbf{ {CIFAR-10}}}&\multicolumn{3}{c||}{  \textbf{CIFAR-100}}&\multicolumn{3}{c}{  \textbf{ImageNet}}\\
			\hline
			\textbf{Acquisition Function} &   \textbf{12.5\%} &   \textbf{25\%} &   \textbf{50\%}&   \textbf{20\%} &   \textbf{40\%} &   \textbf{80\%}& \textbf{20\%} &   \textbf{40\%} &   \textbf{80\%}\\
			\hline
			Random & 87.65& 91.88& 94.30 & 63.96 & 74.18 & 80.65 & 63.51 & 68.54 & 71.66 \\
			Entropy ($\mathcal{H}$)& 89.67 & 94.29 & 96.08 & 65.83 & 76.35 & 81.94 & 64.11 & 69.64 & 72.00 \\
			Mutual Information ($\mathcal{J}$) & 89.46 & 94.34 & \textbf{96.30} & 66.11 & 76.29 & {82.02} & 64.65 & {70.18} & {72.36} \\
			Variation Ratios ($\mathcal{V}$) & 89.86 & 94.42 & 95.76 & 65.28 & 76.27 & \textbf{82.37} & 64.39 & 69.20 & \textbf{72.78} \\
			Error Count ($\mathcal{E}$) & 87.30 & 94.02 & 96.08 & 65.37 & 76.13 & 82.06 & 58.13 & 64.66 & 72.10 \\
			\hline
            Full Dataset (\textbf{100\%}) &\multicolumn{3}{c||}{96.18}&\multicolumn{3}{c||}{81.86}&\multicolumn{3}{c}{72.33}\\
            \hline
    \end{tabular}
	\end{center}
	\vspace{-0.35cm}
\end{table*}
\section{Data Subset Search for Object Detection}

In this section, we describe the key changes required for applying the presented ensemble AL techniques for object detection. We present a new initialization scheme, automatic duplication, suitable for the long-tailed nature of object detection datasets, and discuss acquisition functions applicable to object detectors.

\subsection{Initialization Scheme}
For the automatic duplication initialization scheme, we consider the case where sampling is performed not only over the remaining dataset, but also over the data previously used for training, as shown in Fig. \ref{fig:dup}. In this case, the selection could lead to repeated samples in the training set, potentially beneficial for training the model due to consistent high uncertainty. In practice, this acts as an automatic approach to improve the balance between classes in the training distribution.

\subsection{Acquisition Function}
In this work, we assume the object detector outputs a 2D map of probabilities per class (bicycle, person, car, etc.). Each position in this map corresponds to a patch of pixels in the input image, and the probability specifies whether an object of that class has a bounding box centered there. Such an output map is often found in single-stage object detectors such as SSD \cite{Liu2016Ssd} or YOLO \cite{Redmon2016You}. For this class of models, we compute the acquisition functions as described in Section \ref{sec:acq} by considering each unique location and class combination in the 2D map to be a binary classification output. By doing so, we obtain \textit{uncertainty heatmaps} for each class (examples are shown in Fig. \ref{fig:heatmaps}). To obtain a final acquisition function at the image-level, we choose the maximum value over all classes for the entire 2D map.

\section{Experiments}
In this section, we demonstrate the effectiveness of AL for building data subsets for image classification and object detection. We initially investigate the impact of the initialization schemes discussed in Section \ref{sec:init} and acquisition functions from Section \ref{sec:acq} for classification with the ResNet-18 architecture \cite{He2016Deep}. We then focus on scaling up the ensemble and evaluating the robustness of our subsets to architecture shifts. We experiment with three classification datasets: CIFAR-10 and CIFAR-100 \cite{Krizhevsky2009Learning}, as well as ImageNet \cite{Deng2009Imagenet}. The CIFAR datasets involve object classification tasks over natural images: CIFAR-10 is coarse-grained over 10 classes, and CIFAR-100 is fine-grained over 100 classes. For both tasks, there are 50k training images and 10k validation images of resolution $32 \times 32$, which are balanced in terms of the number of training samples per class. ImageNet consists of 1000 object classes, with annotation available for 1.28 million training images and 50k validation images of resolution $256 \times 256$. This dataset has a slight class imbalance, with 732 to 1300 training images per class.

For our object detection experiments, we use an internal large scale research dataset consisting of 847k and 33K images respectively for training and testing. Each image is annotated with bounding boxes of up to 5 classes: car, pedestrian, bicycle, traffic sign and traffic light. There is significant class imbalance, which we counter by using class-wise loss weighting terms based on the inverse frequency of the bounding boxes of each class in the training dataset. For evaluation on this task, we consider the performance of a single model and report the weighted mean average precision (wMAP) which averages MAP across several object sizes, prioritizing large objects.

\subsection{Implementation Details}
Unless otherwise specified, we use 8 models with the ResNet-18 \cite{He2016Deep} architecture to build the acquisition and subset models for classification, and 6 single-stage object detector models based on a UNet-backbone for object detection. For all three tasks, we do mean-std pre-processing, and augment the labeled dataset on-line with random scaling, crops and horizontal flips.

For ImageNet, each ResNet-18 uses the standard kernel sizes and counts. For CIFAR-10 and CIFAR-100, we use a variant of ResNet-18 as proposed in \cite{He2016Identity}. We use Stochastic Gradient Descent with a learning rate of 0.1 and momentum of 0.9, and weight decay of $10^{-4}$. On CIFAR, we use a patience parameter (set to 25) for counting the number of epochs with no improvement in validation accuracy, in which case the learning rate is dropped by a factor of 0.1. We end training when dropping the learning rate gives no improvement in the validation accuracy after a number of epochs equal to twice the patience parameter. If the early stopping criterion is not met, we train for a maximum of 400 epochs. On ImageNet, we train for a total of 150 epochs, scaling the learning rate by a factor of 0.1 after 70 and 130 epochs. Experiments are run on Tesla V100 GPUs.

\begin{table}[!t]
	\begin{center}
    \caption{Effect of increasing the number of models in an acquisition ensemble trained with 40\% of ImageNet. Results shown are top-1 accuracy evaluated on the 40\% selected data (512k samples used for training) and 60\% unselected data (768k remaining samples). Gains in performance on the unselected data show that large scale checkpoint-based ensembles exhibit high diversity.}
    \vspace{-0.5cm}
    \resizebox{\linewidth}{!}{
    \label{tab:generalize}
		\begin{tabular}{c|c|c|c|c|c}
			\hline
			\textbf{Eval Set} & \textbf{Single (1)} & \textbf{Seeds (5)} & \textbf{Checkpoints (5)} & \textbf{Checkpoints (20)} & \textbf{Combined (100)} \\
			\hline
		    Selected & 58.10 & 74.98 & 72.59 & 79.60 & 83.78 \\
		    Unselected & 70.85 & 82.55 & 81.47 & 84.02 & 85.57 \\
            \hline
        \end{tabular}
        }
	\end{center}
	\vspace{-0.0cm}
\end{table}

\subsection{Classification: Main Experiments}

\subsubsection{Initialization Schemes}

In our first experiment, we compare the three initialization schemes introduced in Section \ref{sec:init} to a random subsampling baseline. For this experiment, we fix the number of ensemble members to 8 for CIFAR and 4 for ImageNet, each with a different random seed. We fix the acquisition function to Mutual Information ($\mathcal{J}$), as defined in Eq. \ref{eqn:mi}. For the pretrain scheme, we pretrain 8 (or 4) models with different random seeds on the entire dataset $L$, and finetune them starting with a learning rate of $10^{-3}$ on the chosen subset $S$. For the other schemes, we train the subset model from scratch on $S$. We report the top-1 validation accuracy of three independent ensembles, each from a different experimental trial, plotting the mean with one standard deviation as an error bar. These results are summarized in Fig. \ref{fig:init}.

We observe certain common trends for all three datasets: random subsampling (blue in Fig. \ref{fig:init}) leads to a steady drop-off in performance; and the pretrain scheme (orange in Fig. \ref{fig:init}) does not significantly impact the performance in comparison to training with the full dataset. Interestingly, the compress scheme (green in Fig. \ref{fig:init}) performs extremely poorly when the subset chosen is very small, but performs well when the acquisition and subset models have a similar overall dataset size (eg., 1M samples on ImageNet). The poor performance of the compress scheme implies that the uncertainty estimates for this experiment are not robust to large changes in the dataset distribution between the acquisition and subset models. The build up scheme (red in Fig. \ref{fig:init}) consistently outperforms random subsampling by a large margin. The performance is robust across all three tasks. Based on these observations, we fix the initialization scheme to build up for the next experiments. With a sufficiently large subset of the data (eg., 25k on CIFAR-10, 1M on ImageNet), the build up scheme slightly outperforms a model trained on the full dataset.

\subsubsection{Acquisition Functions}

In our next experiment, we compare random subsampling against the four acquisition functions from Section \ref{sec:acq}. Similar to the previous experiment, we use ensembles of 8 members for CIFAR and 4 members for ImageNet for both the acquisition and subset models. We run three experimental trials, and report the mean validation accuracy of the subset model ensemble for each trial, in Table \ref{tab:subsample}.

For all four acquisition functions, the subset models significantly outperform the baseline (random) at the final iteration. Further, when using 50\% of the data on CIFAR-10, and 80\% of the data on CIFAR-100 and ImageNet, we obtain subsets of data that improve performance compared to training on the full dataset (100\%). Among the four functions, mutual information ($\mathcal{J}$) and variation ratios ($\mathcal{V}$) outperform entropy ($\mathcal{H}$) and error count ($\mathcal{E}$). Interestingly, though the $\mathcal{E}$ acquisition function is similar to $\mathcal{V}$, and also has access to the ground truth labels, the data selected by it in the first iteration leads to poor performance, which is not completely recovered in the subsequent iterations. This indicates that a sample for which all the ensemble members collectively make an error may be 'too difficult' and therefore not an ideal choice for the training dataset.

Among the two best acquisition functions, on the CIFAR-10 dataset, $\mathcal{J}$ outperforms $\mathcal{V}$. This is because there are very few absolute disagreements in this setting due to the small number of classes and high performance, which leads to a very small number of samples with non-zero values for $\mathcal{V}$. However, on the CIFAR-100 and ImageNet datasets, with more classes and lower overall performance, there is a greater number of disagreements, and $\mathcal{V}$ outperforms $\mathcal{J}$ by a significant amount when using 80\% of the dataset. Compared to training on the full dataset, using $\mathcal{V}$, we obtain a 0.5\% absolute improvement in validation accuracy on both CIFAR-100 and ImageNet. Additionally, this performance improvement is accompanied by a 20\% reduction in training time on both these tasks. 

\begin{table}[!t]
	\begin{center}
    \caption{Effect on the subset model of increasing the number of ensemble members in the acquisition model using the $\mathcal{V}$ acquisition function. Results shown are top-1 validation accuracy of a single ResNet-18 model trained using 80\% of the ImageNet dataset selected with different ensemble configurations, compared to a baseline of random acquisition. Best results are obtained by the combined configuration, which uses different random seeds and training checkpoints in the acquisition model.}
    \label{tab:ensemble}
    \vspace{-0.25cm}
    \resizebox{\linewidth}{!}{
		\begin{tabular}{c|c|c|c|c}
			\hline
			\textbf{Random}  & \textbf{Seeds (5)} & \textbf{Checkpoints (5)} & \textbf{Checkpoints (20)} & \textbf{Combined (100)} \\
			\hline
		    69.24 & 69.97 & 70.10 & 70.18 & 70.34 \\
            \hline
        \end{tabular}
        }
	\end{center}
	\vspace{-0.4cm}
\end{table}

\subsection{Classification: Additional Experiments}

\subsubsection{Ensemble Configurations}
We now explore the possibility of further gains in performance by scaling up the ensemble to increase the number of samples drawn in the Monte Carlo estimator as per Eq. \ref{eqn:uncsamp}. For the remaining experiments, we focus on the final AL iteration for the ImageNet dataset as per the build up scheme. To do this, we start by setting up 5 different training runs on 40\% of the ImageNet dataset (512k samples) as selected by the best performing acquisition function in Table \ref{tab:subsample} ($\mathcal{J}$). For each of these 5 training runs, we store the 21 checkpoints obtained in the final stage of training (epochs 130-150). We pick 4 ensemble configurations from these ResNet-18 training runs to utilize as the acquisition model for an ablation study: (i) random seeds, which uses a total of 5 models from the best performing epoch of each run; (ii) 5 checkpoints, which uses the 5 models from epochs 130:5:150 of a single run; (iii) 20 checkpoints, which uses the 20 models from epochs 131:1:150 of a single run; and (iv) combined, which uses the '20 checkpoints' combined over all 5 runs to give 100 models.

We initially evaluate the performance of these four ensemble configurations on the data available for sampling. To this end, we report the top-1 accuracy of the four ensemble configurations, along with a baseline of a single model, when evaluated on the 40\% selected data (i.e, the training set $S$) and 60\% unselected data (which is remaining in $L$) of ImageNet. Our results are shown in Table \ref{tab:generalize}. Results for the Single (1), Checkpoints (5) and Checkpoints (20) columns use the best seed of 5 runs. As the number of members in the ensemble is scaled up, we observe large and clear improvements in performance on both selected and unselected data. This shows that the checkpoints obtained with no additional computational cost at train time can be used to generate diverse ensembles.

\begin{figure}[t!]
\centering
\includegraphics[width=0.85\columnwidth]{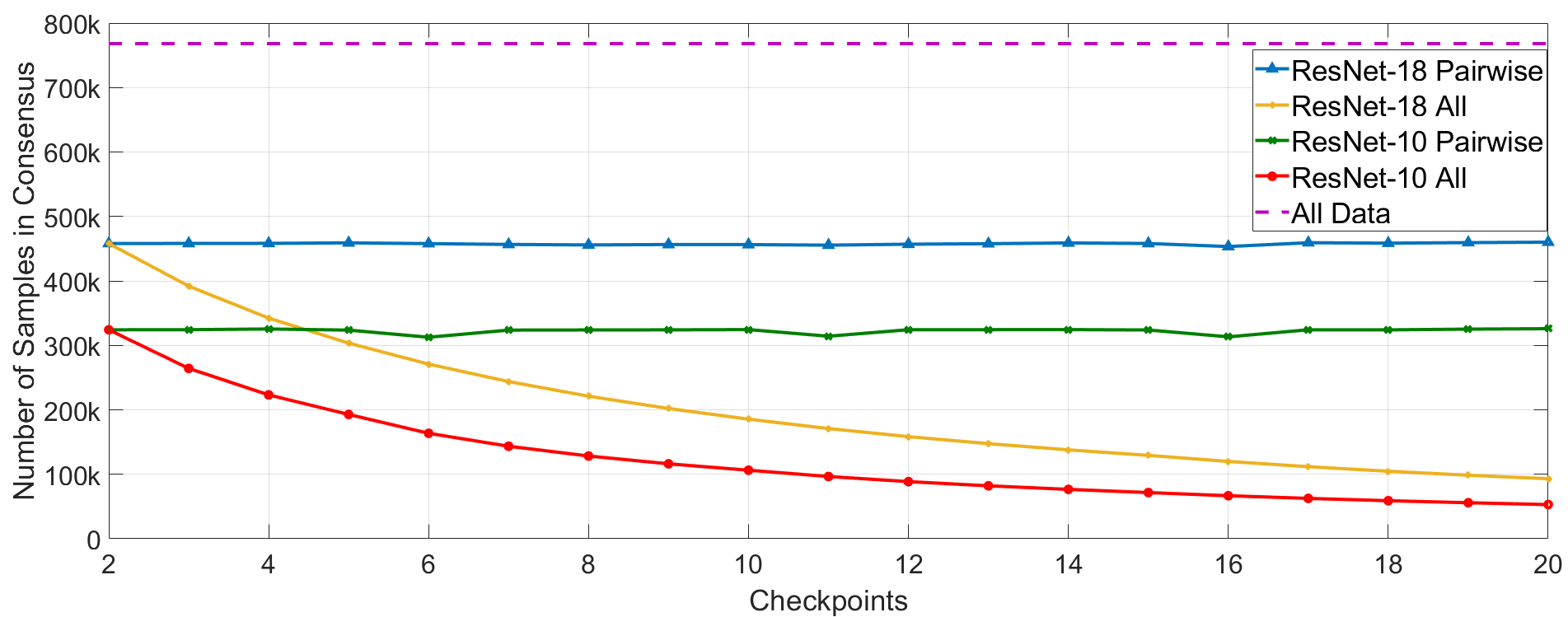}
\vspace{-0.350cm}
\caption{Prediction consensus between different checkpoints during ImageNet training. We observe very little, but consistent agreement in predictions between any consecutive pair of checkpoints. The overall agreement decays rapidly. With 20 checkpoints, there is only consensus on around 90k samples for ResNet-18 and 50k samples for ResNet-10 out of the 768k samples used for this evaluation.}
\label{fig:ckpt}
\vspace{-0.250cm}
\end{figure}

The significant gains (around 3\%) in performance on unselected data in Table \ref{tab:generalize} as the number of models is increased from 5 to 100 indicates the potential for better sampling by scaling up. It is additionally worth noting that for all ensemble configurations, the performance is better on the larger unselected subset of data than on the selected training set. For a single model, the gap in top-1 accuracy between these two subsets is nearly 13\%. This demonstrates the huge amount of redundancy in the unselected part of the dataset as a result of our AL based selection.

Further, we are interested in how the ensemble configurations affect the acquisition function. To this end, we query for an additional 40\% of the unselected data (512k samples) using the variation ratios ($\mathcal{V}$) acquisition function, with each of the four ensemble configurations. We obtain four new subsets, each with 80\% of the samples in ImageNet. The top-1 validation accuracy of a subset model trained using each of these new subsets is shown in Table \ref{tab:ensemble}, along with a baseline of random sampling of 80\% of the data. We observe a steady increase in performance as the number of models in $\mathcal{M}_a$ during acquisition is increased, showing the benefits of scaling up ensemble AL. In particular, the combined configuration improves top-1 accuracy over random sampling by 1.1\%. Note that the earlier results in Table \ref{tab:subsample} use an ensemble of 4 models for evaluation; but Table \ref{tab:ensemble} always evaluates a single model, to allow for a fair comparison between the datasets acquired with 5 models vs. 100 models. Our results show that scaling up is key to exploiting implicit ensembling techniques for AL. Existing work that uses the related idea of snapshot ensembles for AL performs poorly \cite{Beluch2018Power}. This is likely due to (i) the small number of models used, and (ii) the smaller-scale experimental setting (only 2k samples acquired each iteration).

\begin{figure}[t!]
\centering
\includegraphics[width=0.9\columnwidth]{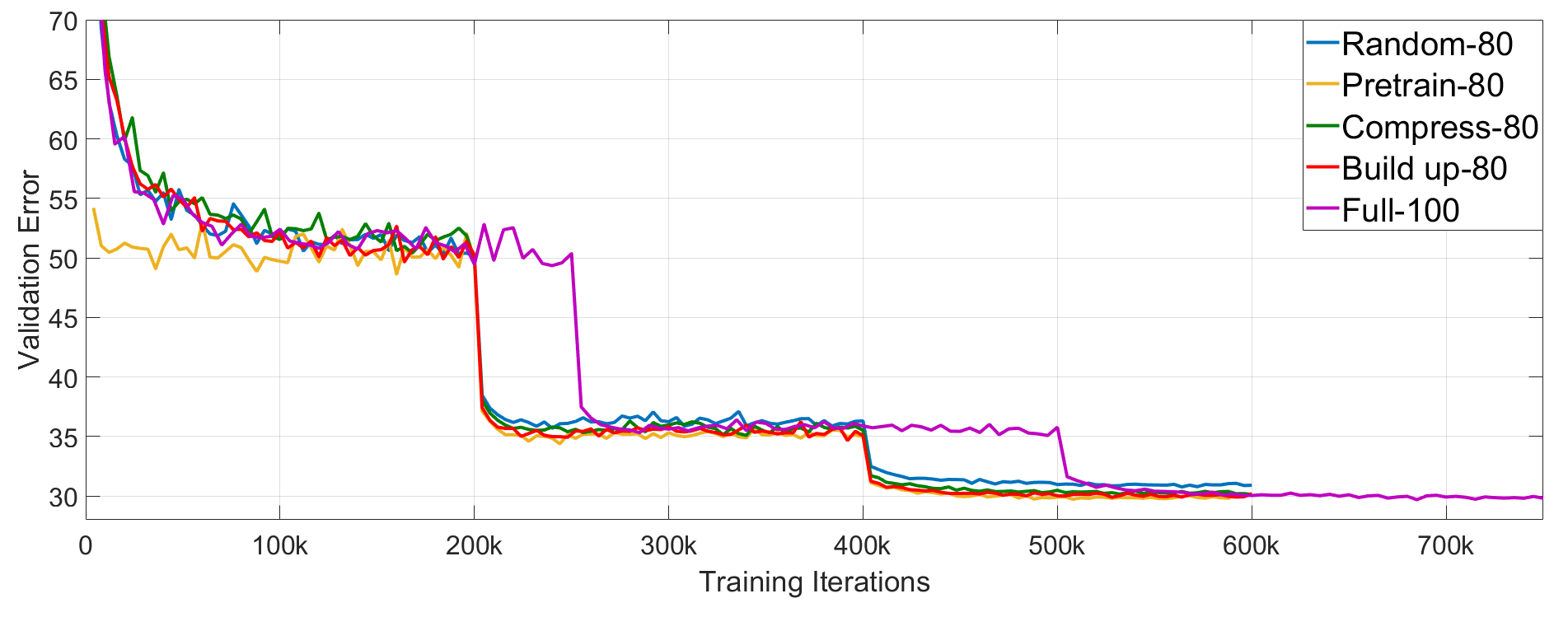}
\vspace{-0.50cm}
\caption{Comparing initialization schemes for ImageNet ResNet-18 training with variation ratios. All three initialization achieve similar performance. Compared to training on the full dataset, convergence time is reduced by 20\%.}
\label{fig:valcurve}
\vspace{-0.0cm}
\end{figure}

\subsubsection{Checkpoint Consensus}
We extend our analysis on the 768k unselected samples of ImageNet from the setting in Table \ref{tab:generalize}, by checking the consensus between the group of $n$ final training checkpoints using both the ResNet-18 and ResNet-10 architectures for $n=2$ to $20$. In addition, we check the consensus in predictions for every consecutive pair of checkpoints as a reference. Our results are presented in Fig. \ref{fig:ckpt}. We observe that for ResNet-18 training, any consecutive pair of checkpoints only agree on the top-1 prediction of around 450k samples (blue in Fig. \ref{fig:ckpt}, which is 58\% of the data). All 20 checkpoints of a single run only agree on 90k samples (orange in Fig. \ref{fig:ckpt}, which is 11.7\% of the data). This is surprising, since the top-1 accuracy of each model on the unselected data is above 70\%, indicating that though they are all from the same training run and have similar accuracy, each checkpoint makes different kinds of errors. These results also provide further support to the findings of \cite{Toneva2018Empirical}, which show that a very small subset of samples are 'unforgettable' once learned by the network, and many samples are repeatedly relearned and forgotten. Since the number of disagreements provides valuable information about the uncertainty of each sample for the acquisition functions used in our study, these results showcase the benefits of using more models in an ensemble for uncertainty estimation. The consensus trends for both pairwise and all models remain similar for ResNet-10, hough the absolute values are lower than ResNet-18 due to the lower model accuracy (green and red in Fig. \ref{fig:ckpt}).

\subsubsection{Validation Curves}
To analyze the runtime and convergence of different models, we plot the top-1 validation error of a single ResNet-18 model for the three initialization schemes from Section \ref{sec:init}. This experiment is run using the variation ratios acquisition function on 80\% of ImageNet in three settings: Pretrain-80, Compress-80 and Build-Up-80. We compare to the baselines of 80\% randomly sampled data (Random-80) and the full dataset (Full-100). Our results are shown in Fig. \ref{fig:valcurve}. Initially, the error for Pretrain-80 is lower than the other approaches, but towards the end of training, Pretrain-80, Compress-80 and Build-Up-80 obtain a similar validation accuracy. All three initialization schemes converge to the best validation error in 20\% less time than using the full dataset.

\begin{table}[!t]
	\begin{center}
    \caption{Effect of leaving out the highest uncertainty samples as outliers while sampling 50\% of the CIFAR-10 and CIFAR-100 datasets with the compress initialization scheme and mutual information acquisition function. Results shown are the mean top-1 validation accuracy over 3 trials. We observe that the highest uncertainty samples are outliers on CIFAR-100, but this is not the case on CIFAR-10.}
    \label{tab:outliers}
    	\vspace{-0.25cm}
		\begin{tabular}{c|c|c|c}
			\hline
			 &\textbf{No Outliers} &  \textbf{12.5\% Outliers} &  \textbf{25\% Outliers} \\
            \hline
			\textbf{CIFAR-10}&\textbf{95.77} & 94.85 & 93.81 \\
            \hline\hline
			\textbf{CIFAR-100}& 75.89 & \textbf{76.76} & 76.41 \\ \hline
        \end{tabular}
	\end{center}
	\vspace{-0.25cm}
\end{table}


\begin{table}[!t]
	\begin{center}
    \caption{Transferring the 80\% subset of ImageNet using ResNet-10 and ResNet-18 to new architectures for the subset model. We observe that these subsets provide near equivalent performance to training on the entire dataset (Full-100) across all new architectures.}
    \label{tab:robust}
    \vspace{-0.35cm}
    \resizebox{\linewidth}{!}{
		\begin{tabular}{c|c|c|c|c|c}
			\hline
			\textbf{Dataset} & \textbf{ResNet-18} & \textbf{ResNet-34} & \textbf{ResNet-50} & \textbf{ResNet-101} & \textbf{DenseNet-121} \\
			\hline
		    Random-80 & 69.24 & 73.00 & 75.15 & 76.72 & 74.59 \\
		    AL-R10-80 & 70.31 & 73.49 & 75.91 & 77.87 & 75.28 \\
		    AL-R18-80 & 70.34 & 73.61 & 76.18 & 77.75 & 75.42 \\
            \hline
            Full-100 & 70.12 & 73.68 & 76.30 & 77.99 & 75.30 \\
            \hline
        \end{tabular}
        }
	\end{center}
	\vspace{-0.0cm}
\end{table}

\subsubsection{Outliers}

It is possible that a very small subset with high uncertainty is informative to the acquisition model, but \textit{too difficult} for the subset model which is trained from scratch with just these samples. To check for this, we repeat the experiment for the compress scheme on CIFAR from Fig. \ref{fig:init} while choosing a subset of 25k samples (50\%), but set aside a percentage of the highest uncertainty samples as outliers instead of adding them to $S$ them during acquisition. For example, for 12.5\% outliers, after sorting by the acquisition function, we select the samples in the range 37.5\% to 87.5\% as the subset $S$ instead of 50\% to 100\%. As shown in Table \ref{tab:outliers}, leaving outliers does not improve the performance of the compress scheme on CIFAR-10. Even in the case of CIFAR-100, where there is an improvement in performance after removing outliers, the obtained accuracy of 76.76\% is well short of the build up scheme (red in Fig. \ref{fig:init}), which reaches 79.37\%. This indicates that though the highest uncertainty samples are sub-optimal for training when taken alone, they are important when used as part of a larger set of samples. 

Additionally, we conduct the same experiment checking for outliers on ImageNet with the build up scheme in the setting of Table \ref{tab:ensemble}. In this experiment, we sample 80\% of the dataset while leaving out 50k samples (approximately 4\%) as outliers for the 'combined' ensemble configuration. These are those samples for which every single model in the ensemble predicted a different class. Doing so reduces subset model performance from 70.34\% to 70.09\%, indicating that the highest uncertainty samples are indeed crucial to outperforming the model trained on the full dataset.



\begin{table}[!t]
\caption{Training data subset search with AL for object detection on an internal benchmark for prototyping perception models for autonomous driving. Just one iteration of AL with either initialization scheme leads to better performance than training with the full dataset on most classes. While the performance of the build up scheme regresses back to the full dataset as more data is added, the automatic duplication scheme continues to improve with iterations.}
    \centering
    \resizebox{\columnwidth}{!}{
    \begin{tabular}{l|p{1cm}|c|c|c|p{1.2cm}}
    \hline
   \textbf{Iteration} &  \textbf{\# Train images} & \textbf{Method}& \textbf{Initialization Scheme} & \textbf{wMAP}&   \textbf{\# Unique images} \\ \hline
\multirow{3}{*}{1}& \multirow{3}{*}{300k} &\multirow{2}{*}{AL} & Automatic Duplication &70.8&277k\\
& & &Build Up &70.9&300k\\ \cline{3-6}
& &Random& - &67.7&300k\\ \hline
\multirow{3}{*}{2}& \multirow{3}{*}{500k}&\multirow{2}{*}{AL} & Automatic Duplication &71.9&355k\\
& & & Build Up&70.5&500k\\ \cline{3-6}
& &Random& -& 68.5&500k\\  \hline
\multirow{3}{*}{3}& \multirow{3}{*}{ 700k }&\multirow{2}{*}{AL} & Automatic Duplication &\textbf{73.2}&402k\\
& & & Build Up&69.5&700k\\ \cline{3-6}
& &Random& - &69.2&700k\\ \hline
 --&~847k & \multicolumn{2}{c|}{Full Dataset}&69.0&847k\\ \hline
\end{tabular}
}
\vspace{-0.0cm}
\label{tab:detection}
\end{table}

\subsubsection{Robustness to Architecture Shift}

Finally, we evaluate the robustness of the subsets to changes in model capacity. Specifically, we evaluate the robustness of the best performing subset selected with a ResNet-18 acquisition model in Table \ref{tab:ensemble} ('combined'). Additionally, we consider the subset obtained with an even more lightweight ResNet-10 model, where we remove 2 convolutional layers from each residual block of a ResNet-18. We use these subsets (referred to as AL-R18-80 and AL-R10-80) to train subset models with the ResNet-18, ResNet-34, ResNet-50, ResNet-101 \cite{He2016Deep} and DenseNet-121 \cite{Huang2016Densely} architectures. For reference, we also evaluate a randomly subsampled 80\% of ImageNet (Random-80) and the full dataset (Full-100). Our results are summarized in Table \ref{tab:robust}. As shown, on all 5 architectures, the subset obtained by AL with ResNet-18 achieves similar performance to training on the full dataset, with a 20\% reduction in overall training time. The AL-R10-80 subset also strongly outperforms the random baseline. This ability to transfer selected subsets to larger architectures has significant implications in domains where training time is crucial, such as MLPerf \cite{MLPerf}.

\subsection{Object Detection Experiments}

In our last experiment, our goal is to compare the build up and proposed automatic duplication initialization scheme for object detection. To this end, we train our object detection ensemble on an initial subset of 100k randomly selected images, and iterate 3 times selecting 200k images in each AL iteration. We use the mutual information acquisition function. Results are presented in comparison to the baseline of random data selection.

Table \ref{tab:detection} shows the summary of our results for this experiment. As a reference, we also include the performance of the model trained using all the data available. We observe that towards the final stages of the data subset search, the random sampling baseline provides little to no benefit despite the addition of significant amounts of data. In contrast, we see that both build up and automatic duplication, which consistently outperform random sampling, also improve accuracy compared to training the model with the entire dataset. Interestingly, automatic duplication leads to the best results with a large reduction in the number of unique images selected for training. For instance, in the third iteration we obtain a 4.2\% absolute improvement compared to training with the full dataset while discarding 53\% of the data (only 402k unique images).

\begin{table}[!t]
	\begin{center}
    \caption{Number of frames with multiple occurrences when using the automatic duplication scheme. At every AL iteration, the number frames with only a single occurrence reduces, and about half the frames that were duplicated at one iteration are duplicated again at the next.}
    \label{tab:duplicates}
	\vspace{-0.25cm}
		\begin{tabular}{c|c|c|c|c|c}
			\hline
			\textbf{Iteration} & \textbf{Unique} & \textbf{1 time} & \textbf{2 times} & \textbf{3 times} & \textbf{4 times} \\
			\hline
		    0 & 100k & 100k & - & - & - \\
            1 & 277k & 254k & 23k & - & - \\
            2 & 355k & 223k & 119k & 13k & - \\
            3 & 402k & 207k & 103k & 83k & 9k \\
            \hline
        \end{tabular}
	\end{center}
	\vspace{-0.4cm}
\end{table}

We further analyze the frames added by the automatic duplication scheme, by checking the number of times they are duplicated. These statistics are presented in Table \ref{tab:duplicates}. At iteration 1, we observe that 23\% of the original frames are duplicated, while the other required frames are added from the remaining dataset. As the iterations progress, we observe fewer and fewer samples that have only a single occurrence in the training set. At iteration 2, the majority of the dataset is samples with 2 copies (119k$\times$2), and at iteration 3, the majority of the dataset is samples with 3 copies (83k$\times$3). Only 9\% of the original samples are duplicated at every single AL iteration, indicating that with sufficient copies in the training data, the approach learns to handle the most difficult samples and focus on other parts of the training dataset.

\section{Discussion and Conclusions}
In this paper, we presented an approach to build data subsets for deep neural networks. Our method uses ensemble Active Learning to estimate the uncertainty of each sample in a dataset, and then chooses only the highest uncertainty samples for training. We provide several key insights with this approach in a large-scale setting:
\begin{itemize}
    \item A \textit{Build Up} scheme using an iterative AL loop is more robust across tasks as opposed to \textit{Pretrain} or \textit{Compress} schemes that begin with the entire dataset.
    \item The choice of acquisition function is not the most critical ingredient of the system, all the tested acquisition functions provide reasonable results.
    \item Scaling up the number of models in an ensemble leads to additional accuracy gains, which can be achieved with minimum computational overhead by re-using training checkpoints as an ensemble.
    \item Datasets obtained using AL can be effectively used for training new models with different network architectures or model capacity.
    \item AL can be used to duplicate important frames in tasks involving imbalanced data distributions, leading to significant performance gains.
\end{itemize}
Our results demonstrate that a training data subset search improves the performance of a DNN on three different image classification benchmarks as well as an internal object detection dataset. With the cost of computation and data storage becoming increasingly important, this study has significant practical implications: by reducing dataset sizes, we can reduce the financial and environmental costs of large-scale training \cite{Strubell2019Energy,Schwartz2019Green}.



\bibliographystyle{IEEEtran}
\bibliography{iclr2020_conference}

\ifCLASSOPTIONcaptionsoff
  \newpage
\fi



%

%

\begin{IEEEbiography}[{\includegraphics[width=1in,height=1.25in,clip,keepaspectratio]{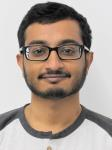}}]{Kashyap Chitta}
is a Ph.D. student with the Autonomous Vision Group led by Prof. Andreas Geiger, part of the Max Planck Institute for Intelligent Systems and University of T\"ubingen, Germany. He obtained his M.S. in Computer Vision in 2018 from Carnegie Mellon University under the supervision of Prof. Martial Hebert. During this time, he also worked as a Deep Learning Intern at NVIDIA on two occasions. 
\end{IEEEbiography}

\begin{IEEEbiography}[{\includegraphics[width=1in,height=1.25in,clip,keepaspectratio]{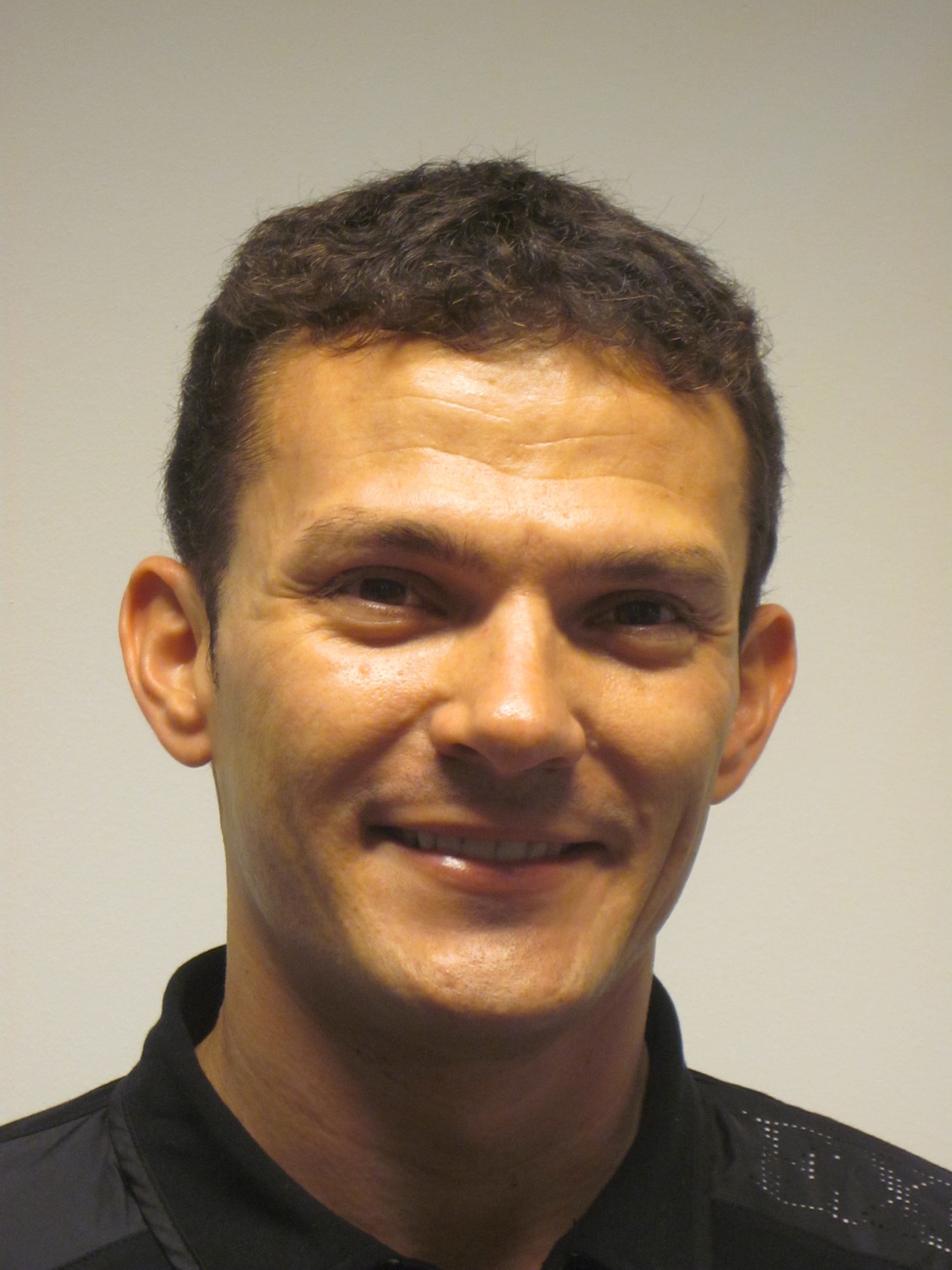}}]{Jos\'{e} M. \'{A}lvarez} is a principal research scientist at NVIDIA. Previously, he was senior researcher at Toyota Research Institute, US; and a senior researcher at Data61, CSIRO, Australia (formerly NICTA). He obtained his Ph.D. from the Autonomous University of Barcelona in 2010 and worked as a postdoctoral researcher at the Courant Institute of Mathematical Science at New York University under the supervision of Prof. Yann LeCun. Since 2014 he is an editor of IEEE-Trans. on Intelligent Transp. Systems.
\end{IEEEbiography}

\begin{IEEEbiography}[{\vspace{-0.6cm}\includegraphics[width=1in,height=1.05in,clip,keepaspectratio]{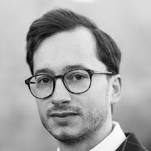}}]{Elmar Haussmann} is a senior deep learning engineer at NVIDIA. Previously, he was the CTO at RiseML, a company he co-founded to cope with the challenges of large-scale machine learning. Elmar completed his Ph.D. studies in 2016 in the area of AI and natural language understanding. Elmar also worked as a software engineer at IBM in the area of high-performance infrastructure.
\end{IEEEbiography}

\begin{IEEEbiography}[{\includegraphics[width=1in,height=1.25in,clip,keepaspectratio]{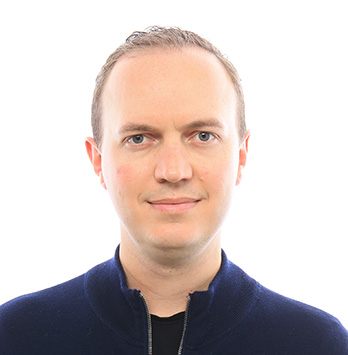}}]{Cl\'{e}ment Farabet} is VP of AI Infrastructure at NVIDIA, developing Maglev, NVIDIA’s platform for self-driving AI development. His focus is to systematize the development of deep neural nets, from principled data selection/expansion/labeling to training and validation. Clement received a PhD from Universit\'{e} Paris-Est in 2013, while at NYU, co-advised by Laurent Najman and Yann LeCun. He co-founded Madbits, a startup focused on web-scale image understanding, sold to Twitter in 2014. Cl\'{e}ment also cofounded Twitter Cortex, a team focused on building Twitter's deep learning platform for recommendations, search, spam, nsfw content and ads.
\end{IEEEbiography}




\clearpage
\appendices

\end{document}